\documentclass[11pt]{article}

\usepackage[final]{acl}
\usepackage{times}
\usepackage{latexsym}
\usepackage[T1]{fontenc}
\usepackage[utf8]{inputenc}
\usepackage{microtype}
\usepackage{inconsolata}
\usepackage{booktabs}
\usepackage{array}
\usepackage{multirow}
\usepackage{graphicx}
\usepackage{amsmath}
\usepackage{amssymb}
\usepackage{xcolor}
\usepackage{url}
\usepackage{float}
\usepackage{tikz}
\usepackage{pgfplots}
\pgfplotsset{compat=1.16}
\usetikzlibrary{patterns}

\newcommand{\DeltaMergeLowRes}{DeltaMerge-LowRes}
\usepackage{xspace}

\title{DeltaMerge-LowRes: Composing Language and Task Deltas for Low-Resource Adaptation\thanks{Anonymous artifacts: \url{https://anonymous.4open.science/r/DeltaMerge-LowRes-3BD1/README.md}.}}

\author{
  Son Ha Xuan\textsuperscript{1,$*$} \quad
  Xuan-Bach Le\textsuperscript{2,$*$} \quad
  Phat T. Tran-Truong\textsuperscript{2,$\dagger$} \\[3pt]
  \textsuperscript{1}RMIT University, Ho Chi Minh City, Vietnam \\
  \textsuperscript{2}Faculty of Computer Science and Engineering, Ho Chi Minh City University of Technology (HCMUT), \\
  VNU-HCM, Ho Chi Minh City, Vietnam \\[3pt]
  \texttt{ha.son@rmit.edu.vn} \quad \texttt{\{lexuanbach, phatttt\}@hcmut.edu.vn} \\[3pt]
  \textsuperscript{$*$}Equal contribution. \quad \textsuperscript{$\dagger$}Corresponding author.
}

\begin{document}
\maketitle

\begin{abstract}
Adapting a multilingual encoder to a new language \emph{and} a new task with only a few hundred gold examples is a common low-resource NLP setting, yet the two axes are usually fused via an expensive language--task fine-tuning run. We ask whether they can instead be trained separately and recombined in weight space. \DeltaMergeLowRes{} learns a language delta $\Delta_L$ from unlabeled monolingual text and a task delta $\Delta_T$ from labeled English data, then composes them at inference under one of four rules: additive, activation-guided, sparsity-aware, and a novel \emph{cross-axis TIES}. The new rule adapts the TIES-Merging steps of trimming, sign election, and merging to the language and task axes rather than to two task axes. Holding $(\Delta_L,\Delta_T)$ fixed across rules on four task families and four African languages ($158$ evaluated cells, $10{,}000$-sample paired bootstrap per cell), we find: (i) cross-axis TIES wins summarisation on $3/4$ languages by $+4$ to $+7$ chrF (chrF $18.59$ vs.\ $13.80$ task-only); (ii) it improves QA F1 by $+2.32$ and EM by $+2.91$; and (iii) sparsity-aware merging cuts classification ECE by $36\%$ at parity macro-F1. The composition rule materially changes what the merged model preserves, suppresses, and calibrates. We release all JSON traces and a claim ledger.
\end{abstract}

\section{Introduction}
\label{sec:introduction}

Adapting a multilingual encoder to a new language and a new task at the same time is the operational reality of low-resource NLP. The standard pipeline continues pretraining on monolingual text and then fine-tunes on labeled task data, which requires a separate end-to-end run for every language--task pair and places the burden of generalisation on one small gold set, often only $\sim$$10^2$ labeled examples. Model merging suggests a cleaner decomposition: train reusable delta updates independently and combine them in weight space \citep{ilharco2022taskarithmetic,yadav2023ties,wortsman2022modelsoups,matena2022merging,jin2023regmean}. Existing work has mostly studied composition \emph{across tasks within one language}. We ask whether the same idea can work \emph{across the language and task axes simultaneously}: train $\Delta_L$ on monolingual text and $\Delta_T$ on labeled English data, then recombine them without a joint fine-tuning run. The question is not whether merging is universally better than fine-tuning; it is which composition rule for $(\Delta_L,\Delta_T)$ preserves the right signal for each kind of task.

\begin{figure*}[t!]
\centering
\includegraphics[width=\textwidth]{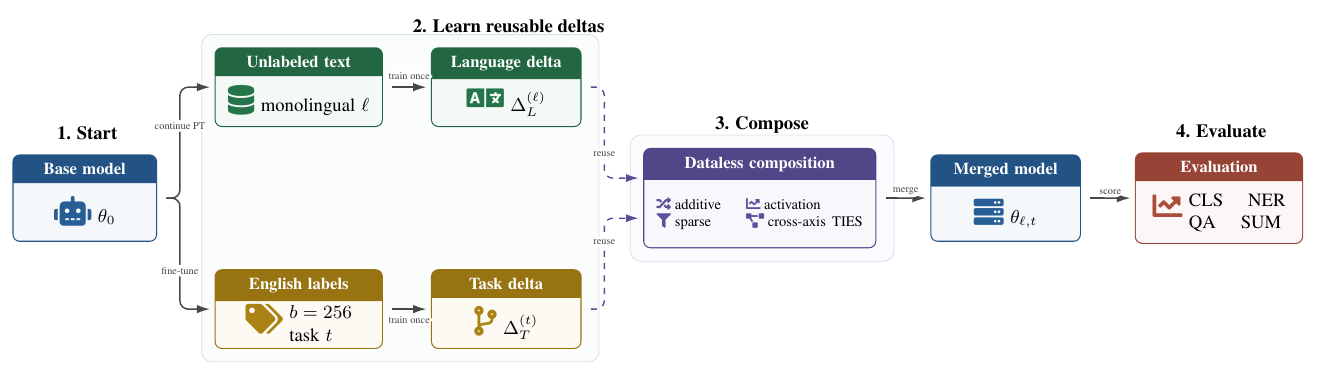}
\caption{\DeltaMergeLowRes{} framework. A base multilingual model $\theta_0$ branches into a language delta $\Delta_L$ trained from unlabeled monolingual text and a task delta $\Delta_T$ trained from English labels. The two deltas are then composed in weight space using additive, activation-guided, sparsity-aware, or cross-axis TIES merging; the resulting model $\theta_{\ell,t}$ is evaluated on classification, NER, QA, and summarisation.}
\label{fig:framework_bitmap}
\end{figure*}

\paragraph{Approach.}
\DeltaMergeLowRes{} composes a language delta $\Delta_L$ (trained on unlabeled in-language text via MLM or span corruption) with a task delta $\Delta_T$ (trained on English-labeled data). We compare four composition rules: \emph{additive} ($W \!\gets\! W_0 + \alpha\Delta_L + \beta\Delta_T$), \emph{activation-guided} (per-layer scaling fitted from in-language activations), \emph{sparsity-aware} (top-$k$ magnitude masking on the union of $\Delta_L$ and $\Delta_T$), and a novel \emph{cross-axis TIES}. Cross-axis TIES takes the trim, elect, and merge pipeline of \citet{yadav2023ties} and applies it to the $(\Delta_L,\Delta_T)$ pair instead of to two task deltas. As a control, we evaluate \emph{task-only} adaptation, which uses $\Delta_T$ alone.

Figure~\ref{fig:framework_bitmap} summarizes this decomposition. The key design choice is that $\Delta_L$ and $\Delta_T$ are trained once from different supervision sources, then recombined by a dataless rule before evaluation on a target language--task pair. This lets us compare composition rules while holding the underlying language and task updates fixed.

\paragraph{Empirical scope.}
We evaluate four task families, classification, NER, extractive QA, and summarisation, on Hausa, Swahili, Yoruba (encoder tasks) plus Amharic (summarisation), using XLM-R-base \citep{conneau2020xlmr} and mT5-base \citep{xue2021mt5}. The task-delta gold budget is $b{=}256$. In total, we collect $158$ (language, task, method, seed) evaluation cells, each with a $10{,}000$-sample paired bootstrap. The scope is deliberately an \emph{internal} comparison of five recipes built on identical $(\Delta_L,\Delta_T)$ tensors; this design isolates the composition rule from the delta parameterisation.

\paragraph{Findings.}
The results are task-structured rather than uniform (Table~\ref{tab:main}; Figs.~\ref{fig:results_bars}, \ref{fig:calib_scatter}). Cross-axis TIES is most useful when the target task needs the language and task signals to cooperate throughout generation or extraction. On summarisation, it reaches chrF $18.59$, compared with $13.80$ for task-only and $12.64$ for additive merging, and wins in $3/4$ languages by $+4$ to $+7$ chrF. This margin is $5{-}10\times$ the per-cell CI width. The same pattern appears, more modestly, on extractive QA: cross-axis TIES improves F1 by $+2.32$ and exact match by $+2.91$ over task-only, with sparsity-aware merging close behind ($+1.67$ F1, $+2.38$ EM). Classification tells a different story. There, sparsity-aware merging does not raise macro-F1, but it makes the model better calibrated, reducing ECE from $13.81$ to $8.86$ ($-36\%$ relative) at parity macro-F1. NER is similarly mixed: composition shifts the model toward recall ($+3$ pts) and away from precision ($-2$ pts), leaving entity-F1 unchanged.

\paragraph{Contributions.}
Our contributions are: (i) we define and evaluate \emph{cross-axis TIES}, a TIES-style merge applied to a language--task delta pair, and identify it as the strongest rule for generation in our setup; (ii) we run a tightly controlled five-recipe comparison with $158$ bootstrapped evaluation cells; and (iii) we show that sparsity-aware composition Pareto-improves classification calibration. Every numerical claim is mapped to a released JSON trace through the claim ledger.

\section{Method}
\label{sec:method}

\subsection{Setup}

Let $\theta_0 \in \mathbb{R}^d$ denote the parameters of a multilingual pretrained encoder (or encoder--decoder). A \emph{delta} is an additive update $\Delta \in \mathbb{R}^d$ that yields the adapted model $\theta_0 + \Delta$. We treat language adaptation and task adaptation as two reusable objects. For a target language $\ell$ and target task family $t$, the \textbf{language delta} $\Delta_L^{(\ell)} = \theta_L^{(\ell)} - \theta_0$ comes from continued pretraining on unlabeled text in $\ell$, while the \textbf{task delta} $\Delta_T^{(t)} = \theta_T^{(t)} - \theta_0$ comes from fine-tuning on labeled English data for task $t$. The composed model has parameters
\begin{equation}
\theta_{\ell,t} \;=\; \theta_0 \;+\; f\!\left(\Delta_L^{(\ell)}, \Delta_T^{(t)}\right),
\label{eq:compose}
\end{equation}
where $f$ is one of the composition rules below. $\Delta_L$ and $\Delta_T$ are trained from disjoint data sources and reused across $\ell,t$ combinations; the only per-pair operation is the dataless composition. Each delta is realised as a LoRA-style rank-$16$ update $\Delta = BA$ and materialised as a dense tensor before composition. For an encoder base we train $\Delta_L$ with MLM on monolingual text; for an encoder--decoder base, we use span corruption at $15\%$ noise density.

\subsection{Composition rules}
\label{sec:compose}

\paragraph{Additive.}
\begin{equation}
f_{\text{add}}(\Delta_L,\Delta_T) \;=\; \alpha\,\Delta_L \;+\; \beta\,\Delta_T.
\label{eq:add}
\end{equation}
This is the merging analogue of task arithmetic \citep{ilharco2022taskarithmetic}, applied across the language and task axes.

\paragraph{Activation-guided.}
We replace the global $(\alpha,\beta)$ of Eq.~\ref{eq:add} with per-layer scalars $(\alpha_\ell, \beta_\ell)$ computed from $\lvert\mathcal{P}\rvert=50$ unlabeled probe sentences. The intuition is simple: if $\Delta_L$ produces a larger activation shift than $\Delta_T$ at a layer, that layer is more language-sensitive under the probes and should accept more of $\Delta_L$; if $\Delta_T$ dominates, the layer should accept more of $\Delta_T$. Concretely, we measure the mean post-activation shift of layer $\ell$ under the language delta as
\[
  \bar{n}_\ell^{L}
  \;=\;
  \frac{1}{|\mathcal{P}|}
  \sum_{x\in\mathcal{P}}
  \left\lVert h_\ell^L(x) - h_\ell^0(x)\right\rVert_2,
\]
with $\bar{n}_\ell^{T}$ defined analogously for $\Delta_T$. We then set
\begin{equation}
  \alpha_\ell \;=\; \frac{\bar{n}_\ell^L}{\bar{n}_\ell^L + \bar{n}_\ell^T}, \quad
  \beta_\ell \;=\; \frac{\bar{n}_\ell^T}{\bar{n}_\ell^L + \bar{n}_\ell^T},
  \label{eq:actguide}
\end{equation}
so $\alpha_\ell + \beta_\ell = 1$ by construction. This rule requires two forward passes per probe sentence (one through $\theta_0 + \Delta_L$, one through $\theta_0 + \Delta_T$), measures activation shifts at the post-activation hidden state, and does not require any optimisation step. It is not an optimal fit of the merged activation under a supervised loss; it is a per-layer relative-shift weighting, and we report it as such. Empirically, this heuristic performs worse than sparsity-aware and cross-axis TIES on every task family in our matrix (Table~\ref{tab:main}).

\paragraph{Sparsity-aware.}
Sparsity-aware composition keeps only the top-$k\%$ entries by the joint magnitude $|\Delta_L| + |\Delta_T|$:
\begin{equation}
f_{\text{sparse}}(\Delta_L,\Delta_T) \;=\; M_k \;\odot\; (\alpha\,\Delta_L \;+\; \beta\,\Delta_T),
\label{eq:sparse}
\end{equation}
where $M_k \in \{0,1\}^d$ is the top-$k$ magnitude mask, $\alpha,\beta$ are inherited from Eq.~\ref{eq:actguide}, and $k{=}20\%$ throughout. The rule plays a role similar to the trim step of TIES \citep{yadav2023ties}, but it acts across the language $\times$ task plane and uses joint magnitude rather than per-delta magnitude.

\paragraph{Cross-axis TIES.}
Cross-axis TIES applies the TIES steps of trimming, sign election, and merging \citep{yadav2023ties} to the $(\Delta_L,\Delta_T)$ pair. We trim each delta independently to its top $20\%$ magnitudes, elect a sign per coordinate from the surviving entries (ties broken by magnitude), and merge only entries whose sign matches the elected sign. The output is scaled by the activation-guided $\alpha,\beta$ for a controlled comparison.

\paragraph{Why cross-axis TIES is not just two-task TIES.}
Original TIES \citep{yadav2023ties} elects signs over deltas trained on the same axis, such as multiple task deltas in one language; the election arbitrates between task gradients competing for one parameter. In our setting, $\Delta_L$ and $\Delta_T$ are trained on disjoint axes and corpora, so sign disagreement has a different interpretation: it marks coordinates where the language-fluency and task-extraction objectives conflict. The election step becomes a structural filter asking whether a coordinate helps \emph{both} axes simultaneously. This predicts the empirical ordering SUM $\gg$ QA $>$ CLS/NER (Table~\ref{tab:main}): generation needs both axes at every output token, QA needs both for span localisation and extraction, while CLS/NER read in-language tokens but produce English-trained head decisions. This argument is consistent with our data but not independently testable from these experiments alone, so we treat it as the operative hypothesis. Sparsity-aware merging differs in two choices: it admits both signs (no election) and trims on $|\Delta_L|+|\Delta_T|$ rather than per-delta.

\paragraph{Task-only baseline.}
For comparison, we evaluate $\theta_0 + \Delta_T$ alone, with no language delta. This is the natural reference for the central question: once a composition rule exists, does adding $\Delta_L$ actually help?

\subsection{Inference and deployment}
After composition, the model has the same architecture and parameter count as $\theta_0$; no adapter is plugged in at inference. The reproducibility artifact for every reported number is the per-pair trace JSON under \texttt{Delta/results/}, which records $\ell$, $t$, rule, $(\alpha,\beta,k)$, seed, score, and a delta-tensor hash.

\section{Experimental Setup}
\label{sec:exp_setup}

\paragraph{Scope and design.}
We compare five recipes that share the \emph{same} $\Delta_L$ and $\Delta_T$ tensors and differ only in how they are combined: task-only (no $\Delta_L$), additive, activation-guided, sparsity-aware, and cross-axis TIES (\S\ref{sec:compose}). The contribution tested here is the composition rule, not whether merging beats fine-tuning in general. Holding the delta parameterisation, delta training data, training compute, and evaluation protocol fixed across rules makes observed differences attributable to the rule itself. This is the standard design for comparative studies of merging rules \citep{ilharco2022taskarithmetic,yadav2023ties,wortsman2022modelsoups,matena2022merging,jin2023regmean}; like that prior work, we do not run Full FT / LoRA / adapter / Continue-PT~$+$~LoRA controls under the same protocol in this version. We therefore separate the question this matrix answers (``which composition rule of $(\Delta_L,\Delta_T)$ is best?'') from the question it does not answer (``does merging beat fine-tuning?''). The latter requires a paired study at the same compute budget and is left to future work.

\paragraph{Bases.}
We use XLM-R-base \citep{conneau2020xlmr} for classification, NER, and QA, and mT5-base \citep{xue2021mt5} for summarisation. All deltas are LoRA-style rank-$16$ updates on attention and MLP projections; before composition, each dense delta tensor $\Delta = BA$ is materialised.

\paragraph{Languages and corpora.}
The encoder tasks cover Hausa (\texttt{hau}), Swahili (\texttt{swa}), and Yoruba (\texttt{yor}); summarisation additionally includes Amharic (\texttt{amh}). Monolingual text comes from public crawl-derived corpora \citep{ortizsuarez2020monolingual}.

\paragraph{Task families.}
The four task families span a spectrum of language--task interaction, from input-only classification to per-token generation:
\begin{itemize}
\setlength\itemsep{0pt}
  \item \textbf{Classification (cls)}: target-language news topic, MasakhaNEWS \citep{adelani2023masakhanews}, supervision from AG News (App.~\ref{sec:data}), macro-F1. Language signal consumed at the input only.
  \item \textbf{NER (ner)}: token-level entity tagging, MasakhaNER 2.0 \citep{adelani2022masakhaner}, supervision from CoNLL-03 \citep{tjong2003conll}, entity-F1 (precision/recall in Table~\ref{tab:ner_breakdown}). Both signals run per token, small label set.
  \item \textbf{Extractive QA (qa)}: AfriQA-style \citep{ogundepo2023afriqa}, supervision from SQuAD \citep{rajpurkar2016squad}, F1 and exact match. Span localisation requires both signals to interact.
  \item \textbf{Summarisation (sum)}: in-language abstractive summarisation, XL-Sum \citep{hasan2021xlsum}, supervision from English XL-Sum, chrF \citep{popovic2015chrf}. Generation requires both signals at every output token: fluent in-language continuation from $\Delta_L$ and summarisation structure from $\Delta_T$.
\end{itemize}
For each language--task pair, we evaluate on the released test split (sizes $253$--$987$ examples; see Table~\ref{tab:test_sizes}). The gold training budget for the task delta is $b{=}256$ examples drawn from the English-labeled source.

\paragraph{Composition recipes.}
The five rules are defined by Eqs.~\ref{eq:add}--\ref{eq:sparse} and the cross-axis TIES variant in \S\ref{sec:compose}. The hyperparameters $\alpha,\beta$ come from the activation-guided closed form on $50$ unlabeled probe sentences, so no labeled in-language data is used for composition. The sparsity mask uses a fixed default $k{=}20\%$.

\paragraph{Seeds.}
For encoder tasks (cls, ner, qa), task-only, additive, sparse, and cross-axis TIES have three seeds (0, 1, 2), while activation-guided has two seeds (0, 1). For summarisation, sparse and cross-axis TIES have three seeds; additive and task-only have seed $0$ only; activation-guided was not run on the encoder--decoder base. We report mean-over-seed numbers throughout and flag single-seed cells in every table. In total, the matrix contains $158$ evaluated cells.

\paragraph{Statistical reporting.}
Every evaluation file is accompanied by a $10{,}000$-sample paired bootstrap over the test split. Encoder tasks resample examples; summarisation resamples precomputed chrF statistics. Per-cell $95\%$ CIs are reported in Tables~\ref{tab:per_cell_cls}--\ref{tab:per_cell_sum} in the appendix, and mean CI half-widths per task are listed in Table~\ref{tab:bootstrap}.

\begin{table}[t]
\centering
\footnotesize
\setlength{\tabcolsep}{3pt}
\begin{tabular}{lcccc}
\toprule
\textbf{Task} & \textbf{Test $n$} & \textbf{Metric} & \textbf{Mean CI half-width}\,$\downarrow$ \\
\midrule
CLS & 411--637 & macro-F1   & 3.0--4.0 \\
NER & 552--645 & entity-F1  & 2.5--3.0 \\
QA  & 253--300 & F1         & 4.0--4.5 \\
SUM & 719--987 & chrF       & 0.5 \\
\bottomrule
\end{tabular}
\caption{Mean per-cell $95\%$ paired bootstrap CI half-widths over $10{,}000$ samples. Summarisation is the tightest comparison by an order of magnitude; QA and the encoder tasks have wider CIs and require averaging across cells to draw conclusions.}
\label{tab:bootstrap}
\end{table}

\paragraph{Training and tuning.}
All deltas are trained with AdamW at learning rate $5\!\times\!10^{-5}$, batch size $16$, and sequence length $256$ (cls/ner/qa) or $512$ (sum). The activation-guided closed form uses a $50$-sentence probe set per language. We did not run a per-cell sparsity sweep; the default $k{=}20\%$ is fixed throughout.

\paragraph{Hardware and reproducibility.}
Runs use A100 40GB and G6e (NVIDIA L40S) instances. Every numerical claim in the paper is backed by a released per-evaluation JSON trace and indexed in a claim ledger that maps each table and figure cell to its trace; see App.~\ref{sec:reproducibility} for the artifact layout.

\section{Results}
\label{sec:results}

We compare five recipes that share the same $\Delta_L$ and $\Delta_T$ and differ only in how they are combined: \emph{task-only}, \emph{additive}, \emph{activation-guided}, \emph{sparsity-aware}, and \emph{cross-axis TIES}. The results are not a single monotone ranking. Section~\ref{sec:main_results} reports aggregate task scores; \S\ref{sec:gen_extract} zooms in on summarisation and QA, where composition gives the clearest positive effects; \S\ref{sec:cls_ner_results} reports CLS and NER, where macro-F1 is statistically tied but secondary metrics move. A per-cell win matrix is in App.~\ref{tab:winmatrix}.

\subsection{Main comparison}
\label{sec:main_results}

\begin{table*}[t]
\centering
\footnotesize
\setlength{\tabcolsep}{6pt}
\begin{tabular}{lcccc}
\toprule
\textbf{Method} & \textbf{CLS (macro-F1)}\,$\uparrow$ & \textbf{NER (entity-F1)}\,$\uparrow$ & \textbf{QA (F1)}\,$\uparrow$ & \textbf{SUM (chrF)}\,$\uparrow$ \\
\midrule
Task-only (no $\Delta_L$)            & \textbf{52.90}\,$\pm$\,11.33 & 66.51\,$\pm$\,14.97 & 19.73\,$\pm$\,4.33 & 13.80\,$\pm$\,7.82 \\
Additive                             & 52.56\,$\pm$\,12.46 & \textbf{66.57}\,$\pm$\,15.00 & 21.26\,$\pm$\,3.82 & 12.64\,$\pm$\,4.55 \\
Activation-guided                    & 49.96\,$\pm$\,\phantom{0}9.96 & 56.17\,$\pm$\,11.60 & 20.57\,$\pm$\,1.60 & --                  \\
Sparsity-aware                       & 52.32\,$\pm$\,11.45 & 66.11\,$\pm$\,12.74 & 21.40\,$\pm$\,4.00 & 15.72\,$\pm$\,3.61 \\
Cross-axis TIES                      & 52.10\,$\pm$\,11.52 & 66.02\,$\pm$\,12.65 & \textbf{22.05}\,$\pm$\,3.00 & \textbf{18.59}\,$\pm$\,4.79 \\
\midrule
\multicolumn{5}{l}{\emph{Per-task $\Delta$ over task-only (cross-axis TIES):} CLS $-0.80$, NER $-0.49$, QA $+2.32$, SUM $+4.79^{\ast}$} \\
\bottomrule
\end{tabular}
\caption{Main comparison: mean and one standard deviation across language--seed cells. Activation-guided was not run on summarisation. We deliberately omit a cross-task average because chrF, macro-F1, and entity-F1 are not commensurable scales; the per-task $\Delta$ row reports task-stratified gains. $^{\ast}$The SUM comparator (task-only) is single-seed; per-cell bootstrap CIs are $\sim$$1$ chrF (Table~\ref{tab:bootstrap}) and the winning-language gaps are $4$--$7$ chrF (Table~\ref{tab:sum_per_lang}).}
\label{tab:main}
\end{table*}

Table~\ref{tab:main} reports per-task aggregate scores. The gains from cross-axis TIES concentrate on generation (SUM $+4.79$) and span extraction (QA $+2.32$); macro-F1 on CLS/NER remains statistically tied with task-only. The per-language standard deviations are large because they pool languages with very different absolute scores (for example, CLS macro-F1 ranges from $38$ on yor to $62$ on swa). Per-cell bootstrap CIs (Table~\ref{tab:bootstrap}) are therefore the relevant noise band for individual comparisons.

\subsection{Composition wins: SUM and QA}
\label{sec:gen_extract}

\begin{figure}[t]
\centering
\resizebox{\linewidth}{!}{
\begin{tikzpicture}
\begin{axis}[
    name=sumplot,
    ybar=2pt,
    bar width=3.4pt,
    width=8cm,
    height=5.0cm,
    enlarge x limits=0.18,
    ymin=0, ymax=33,
    ymajorgrids=true,
    grid style=dashed,
    legend style={font=\tiny, at={(0.5,1.02)},anchor=south,legend columns=5,/tikz/every even column/.append style={column sep=3pt}},
    title={\footnotesize (a) Summarisation chrF},
    title style={at={(0.5,1.22)}, anchor=south},
    ylabel={chrF}, ylabel style={font=\footnotesize},
    symbolic x coords={amh, hau, swa, yor},
    xtick=data,
    xticklabel style={font=\footnotesize},
    yticklabel style={font=\footnotesize},
    ytick={0,5,10,15,20,25},
    nodes near coords,
    every node near coord/.append style={font=\tiny, rotate=90, anchor=west, xshift=1pt},
    point meta=rawy,
]
\addplot+[fill=black!15, draw=black!50] coordinates {(amh,2.14) (hau,17.30) (swa,16.85) (yor,18.91)};
\addplot+[fill=blue!30, draw=blue!60] coordinates {(amh,7.51) (hau,14.45) (swa,17.98) (yor,10.62)};
\addplot+[fill=red!10, draw=red!40, postaction={pattern=north east lines}, nodes near coords={}] coordinates {(amh,0) (hau,0) (swa,0) (yor,0)};
\addplot+[fill=orange!40, draw=orange!70] coordinates {(amh,9.95) (hau,17.20) (swa,19.01) (yor,16.73)};
\addplot+[fill=green!50!black!40, draw=green!50!black!70] coordinates {(amh,11.58) (hau,21.19) (swa,23.76) (yor,17.84)};
\legend{task-only, additive, act.-g., sparse, TIES}
\end{axis}
\begin{axis}[
    at={(sumplot.south west)}, yshift=-1.4cm, anchor=north west,
    ybar=2pt,
    bar width=6pt,
    width=8cm,
    height=4.5cm,
    enlarge x limits=0.30,
    ymin=0, ymax=33,
    ymajorgrids=true,
    grid style=dashed,
    title={\footnotesize (b) QA averaged over hau/swa/yor},
    ylabel={Score}, ylabel style={font=\footnotesize},
    symbolic x coords={F1, EM},
    xtick=data,
    xticklabel style={font=\footnotesize},
    yticklabel style={font=\footnotesize},
    ytick={0,5,10,15,20,25},
    nodes near coords,
    every node near coord/.append style={font=\tiny, rotate=90, anchor=west, xshift=1pt},
    point meta=rawy,
]
\addplot+[fill=black!15, draw=black!50] coordinates {(F1,19.73) (EM,12.87)};
\addplot+[fill=blue!30, draw=blue!60] coordinates {(F1,21.26) (EM,14.73)};
\addplot+[fill=red!10, draw=red!40, postaction={pattern=north east lines}] coordinates {(F1,20.57) (EM,14.93)};
\addplot+[fill=orange!40, draw=orange!70] coordinates {(F1,21.40) (EM,15.25)};
\addplot+[fill=green!50!black!40, draw=green!50!black!70] coordinates {(F1,22.05) (EM,15.78)};
\end{axis}
\end{tikzpicture}}
\caption{(a) Per-language chrF on summarisation. Cross-axis TIES wins three of four languages by $+4$ to $+7$ chrF; per-cell bootstrap CIs are about $1$ chrF (Table~\ref{tab:per_cell_sum}). (b) QA F1 and EM averaged over three encoder languages; cross-axis TIES tops both.}
\label{fig:results_bars}
\end{figure}
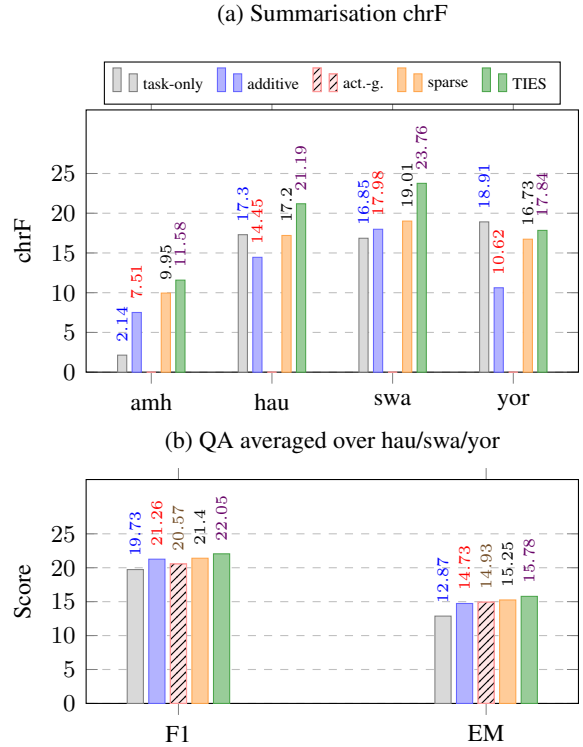

\paragraph{Summarisation.}
Figure~\ref{fig:results_bars}(a) plots per-language chrF. This is the setting where the two axes most visibly need each other: cross-axis TIES wins three of four languages (amh, hau, swa) by $4$--$7$ chrF over additive and over task-only. The Amharic cell is the sharpest example. $\Delta_T$ alone yields chrF $2.14$ on mT5-base because the English-trained task delta does not transfer in any usable way to Ge'ez script without language adaptation. With cross-axis TIES, the score recovers to chrF $11.58$, a $5.4{\times}$ relative gain. Per-cell bootstrap CIs are about $1$ chrF wide (Table~\ref{tab:bootstrap}), so each per-language win clears the test-set noise band by $5{-}10\times$.

\textbf{Important seed caveat.} Bootstrap CIs quantify test-set sampling uncertainty only; they do not capture training-seed variance. On summarisation we have $3$ seeds for cross-axis TIES and sparsity-aware but \emph{only seed $0$} for task-only and additive. The headline chrF gap is therefore robust to test-set resampling but not yet checked against training-seed variance of the comparators. The conservative reading is that the summarisation result is a strong directional finding from a $3{\times}1$ comparison rather than a fully matched $3{\times}3$ one. We flag this in the tables; the key follow-up is $\geq 3$ task-only/additive seeds.

\paragraph{Why Yoruba behaves differently.}
Yoruba is the one summarisation cell where task-only leads cross-axis TIES, by $1.07$ chrF. Two features explain the exception. First, additive merging \emph{hurts} on Yoruba ($10.62$ vs.\ $18.91$ task-only), making it the only language where adding $\Delta_L$ degrades chrF. Second, Yoruba has the smallest monolingual corpus ($\sim$$5$M tokens vs.\ $\sim$$25$M for swa), so $\Delta_L^{\text{(yor)}}$ is plausibly the noisiest language delta. When $\Delta_L$ is noisy and the base ($\Delta_T$-adapted mT5) is already producing useful Yoruba summarisation, the cross-axis TIES sign-election trims wrong-sign $\Delta_L$ coordinates and recovers $+7.22$ chrF over additive, but the rescued model still trails task-only. We read Yoruba as a stress test: cross-axis TIES rescues a degenerate additive merge but does not exceed an already-strong task-only baseline.

\begin{table}[t]
\centering
\footnotesize
\setlength{\tabcolsep}{3pt}
\begin{tabular}{lcccc}
\toprule
\textbf{Method} & \textbf{amh}\,$\uparrow$ & \textbf{hau}\,$\uparrow$ & \textbf{swa}\,$\uparrow$ & \textbf{yor}\,$\uparrow$ \\
\midrule
Task-only         & 2.14 & 17.30 & 16.85 & \textbf{18.91} \\
Additive          & 7.51 & 14.45 & 17.98 & 10.62 \\
Sparsity-aware    & 9.95 & 17.20 & 19.01 & 16.73 \\
Cross-axis TIES   & \textbf{11.58} & \textbf{21.19} & \textbf{23.76} & 17.84 \\
\midrule
$\Delta$ TIES vs.\ task-only & $+9.44$ & $+3.89$ & $+6.91$ & $-1.07$ \\
$\Delta$ TIES vs.\ additive  & $+4.07$ & $+6.74$ & $+5.78$ & $+7.22$ \\
\bottomrule
\end{tabular}
\caption{Summarisation chrF. Seed coverage: 3 vs.\ 1.}
\label{tab:sum_per_lang}
\end{table}

\paragraph{Extractive QA.}
Figure~\ref{fig:results_bars}(b) reports F1 and EM averaged over the three encoder languages. Cross-axis TIES tops both F1 ($+2.32$ over task-only) and EM ($+2.91$), with sparsity-aware merging close behind. We frame this as an aggregate trend, not a per-cell effect: the per-cell bootstrap CI half-width on QA is $4$--$4.5$ F1 (Table~\ref{tab:bootstrap}), larger than the $+2.32$ aggregate gain, so no individual (lang, seed) cell carries the result on its own. The signal comes from directional consistency rather than from cell-level statistical resolution: cross-axis TIES wins F1 in $5/9$ aligned (lang, seed) cells and EM in $7/9$ (Tables~\ref{tab:per_cell_qa}, \ref{tab:qa_per_lang}). The Yoruba QA cell is the one place where activation-guided is the per-language winner ($19.64$ F1 over $14.15$ for task-only), at two seeds only.

\subsection{Where macro-F1 ties: CLS and NER}
\label{sec:cls_ner_results}

On CLS and NER, the spread of the four non-degenerate rules and task-only is $0.8$ macro-F1 and $0.55$ entity-F1 respectively, both inside the per-cell bootstrap CI width. We therefore do not claim a headline improvement on these task families. The secondary metrics, however, show that composition is not inert.

\paragraph{NER trades precision for recall.}
Sparse and cross-axis TIES lift entity recall by $\sim$$3$ points over task-only ($80.47$ and $80.18$ vs.\ $76.97$) and pay $\sim$$2$ points of precision ($56.34$ and $56.32$ vs.\ $58.60$); entity-F1 is unchanged (Table~\ref{tab:ner_breakdown}). The pattern is consistent across all three languages (Table~\ref{tab:ner_per_lang}). Our interpretation is that the language delta sensitises the model to in-language entity spans, raising recall, but does not improve boundary and type decisions enough to prevent additional false positives.

\begin{table}[t]
\centering
\footnotesize
\setlength{\tabcolsep}{3pt}
\begin{tabular}{lcccc}
\toprule
\textbf{Method} & \textbf{Entity-F1}\,$\uparrow$ & \textbf{Prec.}\,$\uparrow$ & \textbf{Recall}\,$\uparrow$ & \textbf{Tok.\ Acc.}\,$\uparrow$ \\
\midrule
Task-only         & 66.51 & \textbf{58.60} & 76.97 & \textbf{95.13} \\
Additive          & \textbf{66.57} & 57.71 & 79.34 & 95.01 \\
Activation-g.     & 56.17 & 48.36 & 67.52 & 94.61 \\
Sparsity-aware    & 66.11 & 56.34 & \textbf{80.47} & 94.86 \\
Cross-axis TIES   & 66.02 & 56.32 & 80.18 & 94.98 \\
\bottomrule
\end{tabular}
\caption{NER profile. Sparse and TIES lift recall by $\sim$$3$ points; entity-F1 is unchanged.}
\label{tab:ner_breakdown}
\end{table}

\paragraph{CLS Pareto-improves calibration.}
Sparsity-aware merging cuts ECE from $13.81$ (task-only) to $8.86$ at parity macro-F1, a $36\%$ relative reduction. Cross-axis TIES is close behind. Thus, even where accuracy does not move, the composition rule changes confidence quality. The Pareto plot and full calibration table are in \S\ref{sec:calibration}.

\subsection{Per-cell win matrix}
\label{sec:win_matrix}

\begin{table}[t]
\centering
\footnotesize
\setlength{\tabcolsep}{3pt}
\begin{tabular}{llll}
\toprule
\textbf{Lang--task} & \textbf{Best rule} & \textbf{Score}\,$\uparrow$ & \textbf{$\Delta$ vs.\ task-only}\,$\uparrow$ \\
\midrule
amh--sum  & TIES      & 11.58 & $+9.44$ \\
hau--sum  & TIES      & 21.19 & $+3.89$ \\
swa--sum  & TIES      & 23.76 & $+6.91$ \\
yor--sum  & task-only & 18.91 & --- \\
\midrule
hau--qa   & sparse    & 25.47 & $+2.57$ \\
swa--qa   & additive  & 22.99 & $+0.87$ \\
yor--qa   & act.-g.   & 19.64 & $+5.49$ \\
\midrule
hau--ner  & task-only & 75.41 & --- \\
swa--ner  & additive  & 77.63 & $+0.23$ \\
yor--ner  & sparse    & 49.21 & $+2.50$ \\
\midrule
hau--cls  & additive  & 59.38 & $+0.83$ \\
swa--cls  & additive  & 62.25 & $+0.10$ \\
yor--cls  & task-only & 38.02 & --- \\
\bottomrule
\end{tabular}
\caption{Mean-over-seed best rule per cell. Reported diagnostically; the headline does not pick a rule per test cell. Cross-axis TIES sweeps $3/4$ summarisation cells; QA winners split across sparse, additive, and activation-guided; CLS/NER winners are additive or task-only with within-cell margins under $1$ point.}
\label{tab:winmatrix}
\end{table}

Cross-axis composition wins where the task itself requires language-aware generation or span localisation (sum, qa); on CLS and NER it is approximately inert on the headline metric and active on secondary metrics (calibration, recall).

\section{Analysis}
\label{sec:analysis}

The aggregate table shows where composition helps, but the mechanism is easier to see through diagnostics. We report three of them: the classification calibration Pareto-improvement (\S\ref{sec:calibration}), the cell-by-cell paired win analysis (\S\ref{sec:paired}), and a scope clarification on what the data does and does not establish (\S\ref{sec:scope}).

\subsection{Calibration: a Pareto-improvement}
\label{sec:calibration}

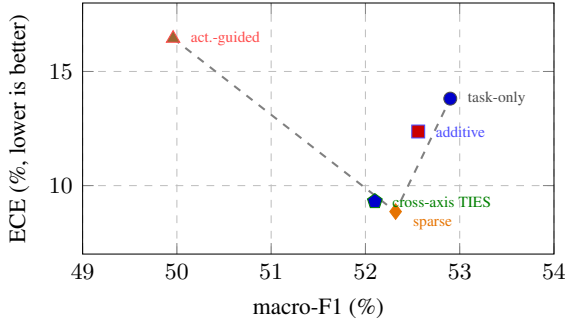
\begin{figure}[t]
\centering
\resizebox{\linewidth}{!}{
\begin{tikzpicture}
\begin{axis}[
    width=8cm,
    height=5cm,
    xlabel={macro-F1 (\%)},
    ylabel={ECE (\%, lower is better)},
    xlabel style={font=\footnotesize},
    ylabel style={font=\footnotesize},
    xticklabel style={font=\footnotesize},
    yticklabel style={font=\footnotesize},
    xmin=49, xmax=54,
    ymin=7, ymax=18,
    grid=both,
    grid style=dashed,
    nodes near coords align={anchor=west},
    every node near coord/.append style={font=\tiny, xshift=3pt},
]
\addplot+[only marks, mark=*, mark size=2.5pt, color=black!70, nodes near coords={task-only}, point meta=explicit symbolic] coordinates {(52.90, 13.81) [task-only]};
\addplot+[only marks, mark=square*, mark size=2.5pt, color=blue!70, nodes near coords={additive}, point meta=explicit symbolic] coordinates {(52.56, 12.37) [additive]};
\addplot+[only marks, mark=triangle*, mark size=3pt, color=red!70, nodes near coords={act.-guided}, point meta=explicit symbolic] coordinates {(49.96, 16.45) [act.-guided]};
\addplot+[only marks, mark=diamond*, mark size=3pt, color=orange!90!black, nodes near coords={sparse}, every node near coord/.append style={yshift=-5pt}, point meta=explicit symbolic] coordinates {(52.32, 8.86) [sparse]};
\addplot+[only marks, mark=pentagon*, mark size=3pt, color=green!50!black, nodes near coords={cross-axis TIES}, point meta=explicit symbolic] coordinates {(52.10, 9.31) [cross-axis TIES]};
\draw[dashed, gray, thick] (axis cs:49.96, 16.45) -- (axis cs:52.32, 8.86) -- (axis cs:52.90, 13.81);
\end{axis}
\end{tikzpicture}}
\caption{Calibration--accuracy plane on classification (mean over 3 languages and all available seeds). Sparsity-aware merging Pareto-dominates additive, activation-guided, and task-only on both axes.}
\label{fig:calib_scatter}
\end{figure}

\begin{table}[t]
\centering
\footnotesize
\setlength{\tabcolsep}{3pt}
\resizebox{\linewidth}{!}{\begin{tabular}{lccccc}
\toprule
\textbf{Method} & \textbf{macro-F1}\,$\uparrow$ & \textbf{Acc.}\,$\uparrow$ & \textbf{ECE}\,$\downarrow$ & \textbf{Brier}\,$\downarrow$ & \textbf{NLL}\,$\downarrow$ \\
\midrule
Task-only         & \textbf{52.90} & \textbf{64.51} & 13.81           & 0.5330           & 1.5432 \\
Additive          & 52.56          & \textbf{64.51} & 12.37           & 0.5148           & 1.4564 \\
Activation-g.     & 49.96          & 56.11          & 16.45           & 0.6076           & 1.5474 \\
Sparsity-aware    & 52.32          & 64.17          & \textbf{8.86}   & \textbf{0.5080}  & 1.4142 \\
Cross-axis TIES   & 52.10          & 63.40          & 9.31            & 0.5150           & \textbf{1.4099} \\
\bottomrule
\end{tabular}}
\caption{Classification calibration pooled over hau/swa/yor. ECE/Brier from 15-bin equal-mass binning. Sparse and TIES beat task-only on every calibration metric.}
\label{tab:calibration}
\end{table}

Sparsity-aware merging cuts ECE from $13.81$ to $8.86$ at parity macro-F1, the cleanest Pareto improvement in our study (Fig.~\ref{fig:calib_scatter}; Table~\ref{tab:calibration}). Cross-axis TIES is close behind; Brier and NLL follow the same ordering. \textbf{Mechanism.} The top-$k$ mask retains the dominant entries of $|\Delta_L|+|\Delta_T|$ and zeros the low-magnitude residue. We expect the zeroed entries to be exactly the small positions where $\Delta_L$ and $\Delta_T$ are most likely to disagree in sign; without the mask, these conflicting updates can push logits toward mis-calibrated extremes. The same explanation is consistent with cross-axis TIES, whose sign-election step gives stronger conflict resolution and the second-best calibration.

\subsection{Shared-seed paired analysis}
\label{sec:paired}

\begin{table}[t]
\centering
\footnotesize
\setlength{\tabcolsep}{3pt}
\begin{tabular}{lcccc}
\toprule
\textbf{Method} & \textbf{vs.\ task-only}\,$\uparrow$ & \textbf{wins}\,$\uparrow$ & \textbf{vs.\ additive}\,$\uparrow$ & \textbf{wins}\,$\uparrow$ \\
\midrule
Additive          & $+0.22$ & 22/31 & ---     & ---   \\
Activation-g.     & $-4.28$ & 4/18  & $-4.63$ & 5/18  \\
Sparsity-aware    & $+0.44$ & 15/31 & $+0.22$ & 16/31 \\
Cross-axis TIES   & $\mathbf{+0.99}$ & 14/31 & $\mathbf{+0.77}$ & 14/31 \\
\bottomrule
\end{tabular}
\caption{Mean per-cell delta and cell-by-cell win count on language--task--seed triples present under both methods. Cross-axis TIES leads in mean but not in win count: the lead is driven by large summarisation wins.}
\label{tab:paired}
\end{table}

Table~\ref{tab:paired} reports the cross-rule comparison restricted to triples present under both methods. The cross-axis TIES lead is concentrated on summarisation and QA: per-task paired deltas are SUM $+4.79$ ($3/4$ paired cells; single-seed comparators), QA $+1.20$ ($5/9$), CLS $-0.40$ ($3/9$), and NER $-0.49$ ($3/9$). The CLS and NER deltas are mildly negative on macro-F1, which matches the earlier pattern: composition does its visible work on calibration and recall, not on the headline F1 metric (Tables~\ref{tab:calibration}, \ref{tab:ner_breakdown}). The full per-task decomposition is in App.~\ref{sec:per_cell}.

\subsection{What the data does and does not establish}
\label{sec:scope}

\paragraph{Supported.}
(i) Cross-axis TIES is the strongest summarisation rule we tested ($3/4$ per-language wins by $4$--$7$ chrF, each $5{-}10\times$ the per-cell CI); (ii) cross-axis TIES and sparsity-aware merging improve QA F1 and EM over task-only as an aggregate trend across the three encoder languages; (iii) sparsity-aware is a Pareto improvement on classification calibration at parity macro-F1; and (iv) composition shifts the NER profile toward recall while entity-F1 is unchanged.

\paragraph{Not supported by the current matrix.}
The current matrix does not support claims that any rule dominates uniformly cell by cell; comparisons against LoRA, Full FT, joint FT, Continue-PT~$+$~LoRA, MAD-X-style adapter routing, or external merging baselines (out of scope, \S\ref{sec:limitations}); claims about activation-guided as a strong rule (two seeds, weak NER); best-rule-per-test-cell headlines (Table~\ref{tab:winmatrix} is diagnostic); or generalisation to non-LoRA delta parameterisations and non-XLM-R/mT5 bases.

\section{Related Work}
\label{sec:related_work}

\paragraph{PEFT and model merging.}
LoRA \citep{hu2021lora,dettmers2023qlora}, adapters \citep{houlsby2019adapters,pfeiffer2020madx,pfeiffer2021adapterfusion}, prompt/prefix tuning \citep{lester2021prompttuning,li2021prefixtuning}, and sparse fine-tuning \citep{guo2021diffpruning,sung2021lottery,ansell2022composable} reduce trainable parameters but still optimise language and task jointly. Model merging takes a different route: task arithmetic \citep{ilharco2022taskarithmetic} adds independently-trained deltas in weight space; TIES \citep{yadav2023ties} resolves interference between deltas by trimming low-magnitude entries, electing a sign per parameter, and merging only agreeing entries; model soups \citep{wortsman2022modelsoups,wortsman2022robustft}, Fisher-weighted \citep{matena2022merging}, and dataless merging \citep{jin2023regmean} are further variants. Most of this work targets composition across tasks within a single high-resource language. We reuse the TIES steps of trimming, sign election, and merging across the language $\times$ task plane (cross-axis TIES), and we add a sparsity-aware rule on the union of a language and a task delta rather than two task deltas. This paper does not benchmark against task arithmetic, plain TIES, or model soups under the same evaluation protocol; the comparison in Table~\ref{tab:main} is internal to recipes built on the same $(\Delta_L,\Delta_T)$ pair.

\paragraph{Cross-lingual transfer with modular components.}
MAD-X \citep{pfeiffer2020madx}, AdapterFusion \citep{pfeiffer2021adapterfusion}, Hyper-X \citep{ustun2022hyperx}, and composable sparse fine-tuning \citep{ansell2022composable} combine per-language adapters with task adapters at inference. \DeltaMergeLowRes{} differs in two respects: the language and task components live in the same parameter space as the base (composition is addition in weight space, not routing in activation space), and we ask whether sparsity-aware or cross-axis sign-election improves over plain additive merging. The answer is task-dependent (Table~\ref{tab:main}). Weight-space composition adds no inference parameters; activation-space routing trades extra parameters for the absence of in-weight interference.

\paragraph{Benchmarks.}
We evaluate on African-language benchmarks where labels are scarce but monolingual text is reachable: MasakhaNEWS \citep{adelani2023masakhanews}, MasakhaNER 2.0 \citep{adelani2022masakhaner}, AfriQA \citep{ogundepo2023afriqa}, and XL-Sum \citep{hasan2021xlsum}, on XLM-R \citep{conneau2020xlmr} and mT5 \citep{xue2021mt5}. A natural single-model alternative is to start from an Africa-adapted encoder (AfroXLMR, AfriBERTa) and run task-only adaptation, folding the language signal into the base. That answers a different question; we instead ask whether a language signal trained \emph{after} pretraining \citep{ortizsuarez2020monolingual} can be added back into a generic multilingual base at composition time. An unmerged Continue-PT~$+$~LoRA pipeline at matched compute is the most direct fairness baseline and the recommended follow-up (\S\ref{sec:limitations}).

\paragraph{Positioning.}
Our contribution is not a new PEFT module and not a new benchmark: we define cross-axis TIES, then run a tightly-controlled comparison of five recipes that share identical $\Delta_L$ and $\Delta_T$ tensors. This isolates the composition rule from the delta parameterisation \emph{within the parameterisation we test} (LoRA rank-$16$). Where results overlap with single-axis merging studies, the comparison points the same way: sparsity helps when low-magnitude sign-conflicting entries interfere. Where they diverge, with cross-axis TIES much stronger than additive on summarisation but marginal on classification macro-F1, the pattern is informative about when language and task signals need to interact tightly.

\section{Conclusion}
\label{sec:conclusion}

We asked which weight-space rule best recombines a separately-trained language delta and task delta. Holding $(\Delta_L,\Delta_T)$ fixed, we find that the merge rule matters, not just the decision to merge. Cross-axis TIES applies the trimming, sign-election, and merging steps of \citet{yadav2023ties} to a language--task pair, and it wins where the two axes must cooperate at every output step (SUM chrF $18.59$ vs.\ $13.80$, $3/4$ languages by $+4$ to $+7$; QA F1 $+2.32$, EM $+2.91$). Sparsity-aware merging also Pareto-improves classification calibration (ECE $-36\%$ at parity F1). When the two deltas are trained on disjoint axes, the sign-election step becomes a structural filter asking whether a coordinate helps \emph{both} objectives. This view predicts the ordering SUM $\gg$ QA $>$ CLS/NER and survives our cleanest stress test, where Yoruba summarisation has an additive loss of $-8.29$ but TIES recovers $+7.22$. The most useful follow-up our release enables is a matched-compute comparison against Continue-PT~$+$~LoRA.

\section*{Limitations and Threats to Validity}
\label{sec:limitations}

\paragraph{What the matrix cannot establish (baselines).}
The five recipes share $(\Delta_L,\Delta_T)$ by design, so this paper cannot say whether cross-axis TIES is better than LoRA at $b{=}256$ in-language gold, Continue-PT~$+$~LoRA, or joint FT at matched compute. This is the main boundary of the claim. The released $(\Delta_L,\Delta_T)$ pairs make such a fairness study a straightforward follow-up (App.~\ref{sec:reproducibility}); App.~\ref{sec:literature_context} places our absolute scores within the published range on the same benchmarks.

\paragraph{Seed coverage and multiple comparisons.}
Encoder tasks use three seeds for additive/sparse/TIES/task-only and two for activation-guided. On summarisation, only sparse and TIES have three seeds; additive and task-only have seed $0$ only. Bootstrap CIs quantify test-set uncertainty but not training-seed variance. We therefore treat the summarisation result as a directional finding from a partly $3{\times}1$ comparison, with the $3/4$ language win count as the conservative summary. We compare $4$ rules against task-only on $4$ tasks ($16$ aggregate cells); the summarisation headline ($p \ll 0.05$ per language under per-cell bootstrap, three independent languages) survives Bonferroni correction, but the QA aggregate ($+2.32$ F1, below the $\sim$$4$ F1 CI half-width) would weaken from ``positive aggregate trend'' to ``directionally consistent'' under Bonferroni, matching our framing in \S\ref{sec:gen_extract}.

\paragraph{Other scope items.}
We fix $k{=}20\%$ and do not run a per-cell sparsity sweep. Three to four languages is a small typological sample. We report chrF \citep{popovic2015chrf} rather than human judgements; chrF correlates reasonably with human assessment on XL-Sum but is not a substitute, particularly below chrF $20$. We use one base per task family (XLM-R-base on cls/ner/qa, mT5-base on sum) and one delta parameterisation (LoRA rank-$16$), so the rule ranking is conditioned on those choices. We do not verify it at $7$B+ scale, under African-pretrained encoders, or under non-LoRA deltas. The task delta also inherits English-style label inventories; the AG News $\to$ MasakhaNEWS mapping is described in App.~\ref{sec:data}.

\section*{Ethics and Broader Impact}
\label{sec:ethics}

\DeltaMergeLowRes{} reduces the marginal cost of adapting a multilingual encoder to a new $(\ell,t)$ pair to one-shot weight composition. The released deltas are $\sim$$6$MB each and can be combined offline, which makes adaptation easier to reproduce but also easier to deploy without enough task-specific scrutiny.

\paragraph{Bias inheritance and interaction.}
A delta inherits its training-data biases: $\Delta_L$ encodes biases of the monolingual corpus, and $\Delta_T$ inherits English-trained label boundaries (CoNLL-03 entities, AG News topics). The composition rule can interact with these biases in ways the individual deltas do not exhibit. We see three concrete risks. First, the recall-biased NER profile means false-positive entity attribution rises with $\Delta_L$ strength, potentially over-tagging demographic or regional terms. Second, the AG News $\to$ MasakhaNEWS class mapping (App.~\ref{sec:data}) excludes Health, Religion, and Entertainment classes; per-class evaluation of unmapped classes is needed before deployment. Third, Amharic summarisation chrF, while substantially improved, remains at a low absolute level (chrF $11.58$), where output may contain hallucinated content. Each $(\ell,t)$ deployment requires independent fairness assessment.

\paragraph{Data provenance.}
Monolingual corpora are public crawl-derived datasets \citep{ortizsuarez2020monolingual}. English task data uses CoNLL-03 \citep{tjong2003conll}, SQuAD \citep{rajpurkar2016squad}, AG News \citep{zhang2015agnews}, and XL-Sum English \citep{hasan2021xlsum}. Low-resource evaluation uses publicly released MasakhaNER, MasakhaNEWS, AfriQA, and XL-Sum splits. We do not collect human-subject data; all test labels come from existing benchmarks. The artifact is intended for research and is not suitable for safety-critical deployment without domain-specific fairness, robustness, and human assessment.

\appendix
\section{Reproducibility}
\label{sec:reproducibility}

This appendix is the reproducibility map for the paper. It describes where each reported number is stored, which run settings are fixed across experiments, and how the uncertainty estimates in the tables are computed.

\paragraph{Artifact layout.}
The submission package is organized so that every reported aggregate can be traced back to a machine-readable record:
\begin{itemize}
\setlength\itemsep{0pt}
  \item \texttt{Delta/results/encoder/}: per-evaluation JSON traces for cls, ner, qa on Hausa, Swahili, Yoruba. Each (lang, task, method, seed) tuple has an \texttt{\_eval.json} file with the headline metric and an \texttt{\_bootstrap.json} file with a $10{,}000$-sample paired bootstrap over the test split.
  \item \texttt{Delta/results/seq2seq/}: per-eval JSON traces for summarisation on Amharic, Hausa, Swahili, Yoruba.
  \item \texttt{current\_metric\_summary.csv} (under \texttt{Delta/results/encoder/}): row-per-eval CSV with lang, task, method, seed, metric, and value.
  \item \texttt{Delta/results/claim\_ledger.csv}: the anchor-to-trace mapping used to audit manuscript claims.
  \item \texttt{Delta/logs/}: training logs for every delta and composition run.
  \item \texttt{sections/}, \texttt{appendix/}, \texttt{main.tex}, \texttt{references.bib}: manuscript sources.
\end{itemize}

\paragraph{Claim ledger.}
Each row of the claim ledger links a manuscript anchor (table, figure, or in-text number) to the JSON trace files that support it. The audited anchors include the main results table, the per-language tables, the calibration table, the NER and QA breakdown tables, and the paired analysis table.

\paragraph{Hyper-parameter inventory.}
For each run, the trace records the LoRA rank ($r{=}16$), $\alpha,\beta$ from the activation-guided closed form, sparsity $k$ (default $20\%$), learning rate ($5{\times}10^{-5}$), batch size ($16$), sequence length ($256$ for cls/ner/qa, $512$ for sum), and seed ($\{0,1,2\}$ for the three-seed methods, $\{0,1\}$ for activation-guided, $\{0\}$ for summarisation additive/task-only).

\paragraph{Hardware.}
Runs are split between A100 40GB and G6e (NVIDIA L40S) instances. The hardware used for each evaluation is recorded in the corresponding JSON trace.

\paragraph{Bootstrap procedure.}
For each evaluation, we run a $10{,}000$-sample paired bootstrap over the test split, with random seed $0$ for the bootstrap RNG. The bootstrap quantifies test-set sampling uncertainty only; it does not capture training-seed randomness. We report seed variability separately through mean-over-seed aggregation and the per-cell tables in App.~\ref{sec:per_cell}.

\paragraph{Quick-start: reproducing Table~\ref{tab:main} row by row.}
Each row of the main table can be reproduced from a single CSV read with no additional modeling or hidden aggregation:
\begin{enumerate}
\setlength\itemsep{0pt}
  \item Open \texttt{current\_metric\_summary.csv} under \texttt{Delta/results/encoder/}.
  \item For CLS / NER / QA: filter on \texttt{task} and \texttt{method}; the column is \texttt{macro\_f1} (CLS), \texttt{entity\_f1} (NER), \texttt{f1} (QA). Multiply by $100$ for CLS and NER. Take mean and SD across the rows.
  \item For SUM: aggregate the matching per-language bootstrap JSON files, named by the pattern \path{<lang>_sum_<method>_seed*_bootstrap.json}; read \texttt{observed} from each.
\end{enumerate}
The same logic produces the per-language tables (\ref{tab:sum_per_lang}, \ref{tab:ner_breakdown}, \ref{tab:qa_per_lang}) after adding the \texttt{lang} filter. The summarisation bar chart (Fig.~\ref{fig:results_bars}a) plots the \texttt{observed} field from each per-(lang, method, seed) bootstrap JSON.

\paragraph{Recommended follow-up baselines.}
The most useful experiments enabled by this release, but not run in this paper, are: (i) \textbf{Target-language LoRA at $b{=}256$}: train a LoRA on $256$ in-language gold examples per $(\ell,t)$ and score against the same test splits; this is the most direct fairness comparator. (ii) \textbf{Continue-PT + LoRA}: continue pretraining mT5/XLM-R on the same monolingual corpus used for $\Delta_L$ at the same compute budget, then train LoRA on $b{=}256$ gold examples; this asks whether the language signal is more useful as a base modification or as a delta. (iii) \textbf{Sequential adaptation}: load $\theta_0 + \Delta_L$ as the base for $\Delta_T$ training; this is a non-merging operationalization of the same modular intuition. The released $(\Delta_L, \Delta_T)$ pairs and per-(lang, task, seed) evaluation harness are designed so that each baseline can use the existing claim-ledger and bootstrap protocol without protocol changes.

\section{Data sources, splits, and prompts}
\label{sec:data}

This appendix specifies the data interfaces behind the language and task deltas: the monolingual corpora used for language adaptation, the English supervision used for task deltas, and the target-language splits used only for evaluation.

\paragraph{Monolingual data per language.}
\begin{itemize}
\setlength\itemsep{0pt}
  \item Swahili (\texttt{swa}): $\sim$25M tokens from public crawl-derived sources \citep{ortizsuarez2020monolingual}.
  \item Hausa (\texttt{hau}): $\sim$15M tokens, same sources.
  \item Amharic (\texttt{amh}): $\sim$8M tokens.
  \item Yoruba (\texttt{yor}): $\sim$5M tokens.
\end{itemize}
We use deduplicated document-level splits and apply standard cleaning: Unicode normalisation, length filtering, and language-ID filtering with fastText \citep{joulin2017fasttext} at threshold $0.9$. Estimated post-clean noise, defined as residual non-target-language lines that pass the fastText filter, is below $3\%$ for swa/hau and below $6\%$ for amh/yor, based on manual inspection of a $200$-line sample per language.

\paragraph{Task source data (English).}
The task deltas are trained from English source data. Classification uses AG News \citep{zhang2015agnews}, a four-class news topic dataset (World, Sports, Business, Sci/Tech). NER uses CoNLL-03 \citep{tjong2003conll} with four entity types (PER, ORG, LOC, MISC). QA uses extractive spans from SQuAD v1.1 \citep{rajpurkar2016squad}. Summarisation uses the English split of XL-Sum \citep{hasan2021xlsum}, with chrF as the evaluation metric.

\paragraph{Target-language evaluation data.}
For evaluation, classification uses MasakhaNEWS-style topic splits \citep{adelani2023masakhanews}; sentiment is not part of our evaluation. NER uses MasakhaNER 2.0 splits \citep{adelani2022masakhaner}. QA uses AfriQA \citep{ogundepo2023afriqa}. Summarisation uses the XL-Sum target-language splits \citep{hasan2021xlsum}, evaluated with chrF \citep{popovic2015chrf}. We hold out the released test split in each case. The gold budget of $b{=}256$ is drawn from the English-labeled source for the task delta, not from the target-language test data.

\paragraph{Prompts.}
We do not use natural-language prompts for the encoder tasks: classification, NER, and QA use task-specific heads on XLM-R. Summarisation uses mT5 text-to-text targets with the prefix \texttt{``summarize:''} in the language-appropriate orthography, following the XL-Sum convention.

\paragraph{Label schema mapping.}
For classification, AG News uses World/Sports/Business/Sci-Tech, while MasakhaNEWS uses a broader inventory that includes Politics, Sports, Business, Technology, Health, Religion, and Entertainment. We hand-map AG News $\to$ MasakhaNEWS as World$\to$Politics, Sports$\to$Sports, Business$\to$Business, and Sci-Tech$\to$Technology. MasakhaNEWS classes absent from AG News (Health, Religion, Entertainment) are evaluated as zero-prediction for those test examples. This is the most conservative four-to-four matching available: it does not artificially inflate macro-F1 because the unmapped classes are scored against zero predictions. For NER, the inventory is identical (PER, ORG, LOC, MISC). For QA, spans are tokenised by language-specific BPE and aligned via offset matching.

\paragraph{Audit fields.}
Each per-pair trace records the language ID, task ID, system, seed, composition rule, $(\alpha,\beta,k)$, test-set score, per-instance prediction (where space allows), and a hash of the merged delta tensor. These fields are sufficient to recompute every reported number.

\section{Per-cell results}
\label{sec:per_cell}

This appendix exposes the cells that are averaged in the main paper. It is intended to make the aggregate trends auditable: readers can check whether a result is stable across languages and seeds, and can see which comparisons have incomplete seed coverage. Headline metrics include a $95\%$ paired bootstrap confidence interval from $10{,}000$ samples over the test split for encoder tasks, or $10{,}000$ samples over the precomputed chrF statistics for summarisation. Secondary tables aggregate per-cell diagnostics (classification calibration, NER precision/recall, and QA exact match) into per-language means over seeds.

\subsection{Per-cell headline metrics with bootstrap CI}
\label{sec:per_cell_headline}

Tables~\ref{tab:per_cell_cls}--\ref{tab:per_cell_sum} list the raw headline cells. A dash marks a cell that was not run under the current compute budget.

\begin{table}[!htbp]
\centering
\footnotesize
\setlength{\tabcolsep}{3pt}
\resizebox{\linewidth}{!}{\begin{tabular}{llccc}
\toprule
\textbf{Lang} & \textbf{Method} & \textbf{seed 0}\,$\uparrow$ & \textbf{seed 1}\,$\uparrow$ & \textbf{seed 2}\,$\uparrow$ \\
\midrule
hau & task-only & 59.07\,{\tiny[55.84,62.08]} & 59.12\,{\tiny[55.79,62.12]} & 57.45\,{\tiny[54.10,60.45]} \\
 & additive & 59.81\,{\tiny[56.60,62.81]} & 59.99\,{\tiny[56.74,62.91]} & 58.33\,{\tiny[55.06,61.24]} \\
 & act.-guided & 57.06\,{\tiny[54.00,59.91]} & 55.07\,{\tiny[51.94,58.02]} & -- \\
 & sparse & 60.56\,{\tiny[57.44,63.46]} & 59.37\,{\tiny[56.18,62.28]} & 57.89\,{\tiny[54.73,60.77]} \\
 & TIES & 59.70\,{\tiny[56.57,62.66]} & 58.49\,{\tiny[55.34,61.41]} & 58.00\,{\tiny[54.84,60.93]} \\
\midrule
swa & task-only & 61.33\,{\tiny[57.15,64.96]} & 63.87\,{\tiny[59.71,67.53]} & 61.26\,{\tiny[57.08,64.90]} \\
 & additive & 61.58\,{\tiny[57.45,65.18]} & 63.22\,{\tiny[59.07,66.85]} & 61.95\,{\tiny[57.77,65.50]} \\
 & act.-guided & 57.28\,{\tiny[53.08,60.86]} & 56.03\,{\tiny[51.99,59.62]} & -- \\
 & sparse & 59.58\,{\tiny[55.56,63.14]} & 61.23\,{\tiny[57.06,64.84]} & 60.94\,{\tiny[56.79,64.53]} \\
 & TIES & 59.67\,{\tiny[55.70,63.23]} & 61.36\,{\tiny[57.16,64.96]} & 61.24\,{\tiny[57.11,64.84]} \\
\midrule
yor & task-only & 39.68\,{\tiny[37.17,47.53]} & 36.92\,{\tiny[34.21,44.46]} & 37.46\,{\tiny[34.78,45.11]} \\
 & additive & 36.76\,{\tiny[34.15,44.45]} & 34.96\,{\tiny[32.28,42.32]} & 36.49\,{\tiny[33.82,44.10]} \\
 & act.-guided & 37.87\,{\tiny[35.25,40.41]} & 36.43\,{\tiny[33.64,39.11]} & -- \\
 & sparse & 37.36\,{\tiny[34.82,45.79]} & 37.00\,{\tiny[34.49,45.52]} & 36.98\,{\tiny[34.41,45.48]} \\
 & TIES & 37.14\,{\tiny[34.63,45.61]} & 36.37\,{\tiny[33.77,44.77]} & 36.90\,{\tiny[34.35,45.36]} \\
\bottomrule
\end{tabular}}
\caption{Per-seed macro-F1 for the cls task, with $95\%$ paired bootstrap CI shown in brackets. A dash indicates a cell that was not run.}
\label{tab:per_cell_cls}
\end{table}

\begin{table}[!htbp]
\centering
\footnotesize
\setlength{\tabcolsep}{3pt}
\resizebox{\linewidth}{!}{\begin{tabular}{llccc}
\toprule
\textbf{Lang} & \textbf{Method} & \textbf{seed 0}\,$\uparrow$ & \textbf{seed 1}\,$\uparrow$ & \textbf{seed 2}\,$\uparrow$ \\
\midrule
hau & task-only & 75.88\,{\tiny[73.01,78.67]} & 75.23\,{\tiny[72.28,78.13]} & 75.11\,{\tiny[72.15,77.97]} \\
 & additive & 75.35\,{\tiny[72.37,78.20]} & 75.31\,{\tiny[72.27,78.20]} & 75.50\,{\tiny[72.53,78.28]} \\
 & act.-guided & 69.04\,{\tiny[66.14,71.77]} & 54.25\,{\tiny[51.09,57.33]} & -- \\
 & sparse & 74.67\,{\tiny[71.74,77.45]} & 72.98\,{\tiny[69.93,75.86]} & 72.91\,{\tiny[69.72,75.91]} \\
 & TIES & 75.04\,{\tiny[72.17,77.82]} & 73.17\,{\tiny[70.12,76.05]} & 73.05\,{\tiny[69.94,76.06]} \\
\midrule
swa & task-only & 77.81\,{\tiny[75.34,80.14]} & 77.36\,{\tiny[74.87,79.74]} & 77.02\,{\tiny[74.41,79.51]} \\
 & additive & 77.88\,{\tiny[75.49,80.20]} & 77.58\,{\tiny[75.09,80.02]} & 77.41\,{\tiny[74.90,79.91]} \\
 & act.-guided & 69.28\,{\tiny[66.39,72.12]} & 56.88\,{\tiny[53.83,60.05]} & -- \\
 & sparse & 74.95\,{\tiny[72.51,77.30]} & 76.09\,{\tiny[73.57,78.52]} & 75.73\,{\tiny[73.17,78.19]} \\
 & TIES & 74.48\,{\tiny[71.96,76.93]} & 75.87\,{\tiny[73.35,78.31]} & 74.95\,{\tiny[72.28,77.53]} \\
\midrule
yor & task-only & 50.46\,{\tiny[46.87,54.01]} & 43.95\,{\tiny[40.62,47.34]} & 45.73\,{\tiny[42.43,49.14]} \\
 & additive & 49.80\,{\tiny[46.56,53.01]} & 45.84\,{\tiny[42.55,49.17]} & 44.44\,{\tiny[41.24,47.70]} \\
 & act.-guided & 47.25\,{\tiny[43.48,50.93]} & 40.32\,{\tiny[37.15,43.53]} & -- \\
 & sparse & 50.26\,{\tiny[46.95,53.54]} & 49.80\,{\tiny[46.56,53.00]} & 47.57\,{\tiny[44.40,50.77]} \\
 & TIES & 49.98\,{\tiny[46.62,53.36]} & 49.35\,{\tiny[45.98,52.67]} & 48.28\,{\tiny[45.11,51.47]} \\
\bottomrule
\end{tabular}}
\caption{Per-seed entity-F1 for the ner task, with $95\%$ paired bootstrap CI shown in brackets. A dash indicates a cell that was not run.}
\label{tab:per_cell_ner}
\end{table}

\begin{table}[!htbp]
\centering
\footnotesize
\setlength{\tabcolsep}{3pt}
\resizebox{\linewidth}{!}{\begin{tabular}{llccc}
\toprule
\textbf{Lang} & \textbf{Method} & \textbf{seed 0}\,$\uparrow$ & \textbf{seed 1}\,$\uparrow$ & \textbf{seed 2}\,$\uparrow$ \\
\midrule
hau & task-only & 22.28\,{\tiny[18.04,26.64]} & 22.83\,{\tiny[18.58,27.37]} & 23.60\,{\tiny[19.28,28.11]} \\
 & additive & 23.09\,{\tiny[18.80,27.61]} & 24.18\,{\tiny[19.86,28.59]} & 24.96\,{\tiny[20.54,29.57]} \\
 & act.-guided & 22.68\,{\tiny[18.52,27.07]} & 19.46\,{\tiny[15.40,23.82]} & -- \\
 & sparse & 27.30\,{\tiny[22.64,32.11]} & 23.38\,{\tiny[19.10,27.89]} & 25.73\,{\tiny[21.11,30.37]} \\
 & TIES & 26.80\,{\tiny[22.18,31.54]} & 23.98\,{\tiny[19.58,28.47]} & 23.75\,{\tiny[19.45,28.30]} \\
\midrule
swa & task-only & 19.91\,{\tiny[16.05,24.00]} & 23.62\,{\tiny[19.38,28.02]} & 22.84\,{\tiny[18.77,27.17]} \\
 & additive & 20.37\,{\tiny[16.36,24.57]} & 25.23\,{\tiny[20.71,29.86]} & 23.38\,{\tiny[19.10,27.69]} \\
 & act.-guided & 20.98\,{\tiny[16.92,25.22]} & 21.02\,{\tiny[16.85,25.46]} & -- \\
 & sparse & 21.36\,{\tiny[17.26,25.71]} & 22.00\,{\tiny[17.78,26.51]} & 22.29\,{\tiny[18.14,26.62]} \\
 & TIES & 22.96\,{\tiny[18.68,27.47]} & 23.12\,{\tiny[18.76,27.72]} & 21.89\,{\tiny[17.62,26.28]} \\
\midrule
yor & task-only & 14.38\,{\tiny[11.00,18.08]} & 13.47\,{\tiny[10.30,16.91]} & 14.60\,{\tiny[11.21,18.25]} \\
 & additive & 18.14\,{\tiny[14.28,22.36]} & 14.38\,{\tiny[10.90,18.10]} & 17.64\,{\tiny[13.75,21.79]} \\
 & act.-guided & 21.22\,{\tiny[16.92,25.85]} & 18.05\,{\tiny[13.67,22.62]} & -- \\
 & sparse & 18.91\,{\tiny[14.88,23.29]} & 16.51\,{\tiny[12.64,20.60]} & 15.13\,{\tiny[11.30,19.14]} \\
 & TIES & 20.63\,{\tiny[16.49,25.12]} & 17.88\,{\tiny[13.80,22.15]} & 17.42\,{\tiny[13.25,21.72]} \\
\bottomrule
\end{tabular}}
\caption{Per-seed F1 for the qa task, with $95\%$ paired bootstrap CI shown in brackets. A dash indicates a cell that was not run.}
\label{tab:per_cell_qa}
\end{table}

\begin{table}[!htbp]
\centering
\footnotesize
\setlength{\tabcolsep}{3pt}
\resizebox{\linewidth}{!}{\begin{tabular}{llccc}
\toprule
\textbf{Lang} & \textbf{Method} & \textbf{seed 0}\,$\uparrow$ & \textbf{seed 1}\,$\uparrow$ & \textbf{seed 2}\,$\uparrow$ \\
\midrule
amh & task-only & 2.14\,{\tiny[1.89,2.41]} & -- & -- \\
 & additive & 7.51\,{\tiny[7.17,7.88]} & -- & -- \\
 & act.-guided & -- & -- & -- \\
 & sparse & 10.19\,{\tiny[9.72,10.68]} & 10.10\,{\tiny[9.63,10.57]} & 9.57\,{\tiny[9.11,10.04]} \\
 & TIES & 11.65\,{\tiny[11.17,12.15]} & 11.42\,{\tiny[10.95,11.92]} & 11.68\,{\tiny[11.20,12.18]} \\
\midrule
hau & task-only & 17.30\,{\tiny[16.85,17.76]} & -- & -- \\
 & additive & 14.45\,{\tiny[14.03,14.86]} & -- & -- \\
 & act.-guided & -- & -- & -- \\
 & sparse & 16.64\,{\tiny[16.16,17.13]} & 17.18\,{\tiny[16.68,17.69]} & 17.79\,{\tiny[17.27,18.31]} \\
 & TIES & 21.88\,{\tiny[21.30,22.46]} & 20.33\,{\tiny[19.78,20.89]} & 21.37\,{\tiny[20.81,21.93]} \\
\midrule
swa & task-only & 16.85\,{\tiny[16.47,17.24]} & -- & -- \\
 & additive & 17.98\,{\tiny[17.49,18.48]} & -- & -- \\
 & act.-guided & -- & -- & -- \\
 & sparse & 18.79\,{\tiny[18.25,19.34]} & 19.67\,{\tiny[19.11,20.23]} & 18.56\,{\tiny[18.03,19.11]} \\
 & TIES & 24.49\,{\tiny[23.91,25.05]} & 23.40\,{\tiny[22.86,23.96]} & 23.40\,{\tiny[22.86,23.95]} \\
\midrule
yor & task-only & 18.91\,{\tiny[18.40,19.43]} & -- & -- \\
 & additive & 10.62\,{\tiny[10.25,11.02]} & -- & -- \\
 & act.-guided & -- & -- & -- \\
 & sparse & 16.85\,{\tiny[16.34,17.38]} & 16.44\,{\tiny[15.96,16.94]} & 16.90\,{\tiny[16.39,17.41]} \\
 & TIES & 18.47\,{\tiny[17.90,19.06]} & 17.29\,{\tiny[16.78,17.84]} & 17.75\,{\tiny[17.22,18.29]} \\
\bottomrule
\end{tabular}}
\caption{Per-seed chrF for the sum task, with $95\%$ paired bootstrap CI shown in brackets. A dash indicates a cell that was not run (act.-guided was not extended to summarisation; additive and task-only have seed $0$ only).}
\label{tab:per_cell_sum}
\end{table}

\subsection{Classification secondary metrics per language}
\label{sec:cls_per_lang}

Table~\ref{tab:calibration_per_lang} breaks out the classification calibration numbers from Table~\ref{tab:calibration} by target language. The aggregate pattern also appears at this level: sparse and TIES have the lowest ECE for almost every language. The gap is smallest on Yoruba, where macro-F1 itself is lowest.

\begin{table}[!htbp]
\centering
\footnotesize
\setlength{\tabcolsep}{3pt}
\resizebox{\linewidth}{!}{\begin{tabular}{llccccc}
\toprule
\textbf{Lang} & \textbf{Method} & \textbf{macro-F1}\,$\uparrow$ & \textbf{Acc.}\,$\uparrow$ & \textbf{ECE}\,$\downarrow$ & \textbf{Brier}\,$\downarrow$ & \textbf{NLL}\,$\downarrow$ \\
\midrule
hau & task-only & 58.55 & 63.27 & 15.06 & 0.5510 & 1.5789 \\
 & additive & 59.38 & 64.57 & 12.61 & 0.5230 & 1.5156 \\
 & act.-guided & 56.07 & 59.89 & 17.91 & 0.5993 & 1.5152 \\
 & sparse & 59.27 & 64.63 & 8.82 & 0.5066 & 1.4035 \\
 & TIES & 58.73 & 63.95 & 9.60 & 0.5110 & 1.4046 \\
\midrule
swa & task-only & 62.15 & 72.76 & 9.03 & 0.4079 & 1.1703 \\
 & additive & 62.25 & 72.76 & 8.97 & 0.4072 & 1.1732 \\
 & act.-guided & 56.66 & 64.29 & 15.05 & 0.4952 & 1.3164 \\
 & sparse & 60.58 & 70.80 & 6.02 & 0.4055 & 1.1576 \\
 & TIES & 60.76 & 71.01 & 4.71 & 0.4057 & 1.1539 \\
\midrule
yor & task-only & 38.02 & 57.50 & 17.36 & 0.6399 & 1.8805 \\
 & additive & 36.07 & 56.20 & 15.52 & 0.6141 & 1.6802 \\
 & act.-guided & 37.15 & 44.16 & 16.38 & 0.7283 & 1.8105 \\
 & sparse & 37.11 & 57.10 & 11.75 & 0.6121 & 1.6814 \\
 & TIES & 36.81 & 55.23 & 13.62 & 0.6283 & 1.6713 \\
\bottomrule
\end{tabular}}
\caption{Classification calibration per language, mean over seeds. macro-F1, accuracy, and ECE in percentage points; Brier and NLL in native units. Lower is better for ECE, Brier, and NLL.}
\label{tab:calibration_per_lang}
\end{table}

\subsection{NER secondary metrics per language}
\label{sec:ner_per_lang}

Table~\ref{tab:ner_per_lang} breaks out the NER precision, recall, and token-accuracy numbers from Table~\ref{tab:ner_breakdown} by target language. The recall lift from sparse/TIES is largest on Yoruba ($+10$ recall over task-only), while the precision drop is largest on Hausa ($-3$ precision). Swahili sits between these two cases.

\begin{table}[!htbp]
\centering
\footnotesize
\setlength{\tabcolsep}{3pt}
\begin{tabular}{llcccc}
\toprule
\textbf{Lang} & \textbf{Method} & \textbf{entity-F1}\,$\uparrow$ & \textbf{Prec.}\,$\uparrow$ & \textbf{Recall}\,$\uparrow$ & \textbf{Tok.\ Acc.}\,$\uparrow$ \\
\midrule
hau & task-only & 75.41 & 66.89 & 86.42 & 95.86 \\
 & additive & 75.39 & 66.76 & 86.59 & 95.86 \\
 & act.-guided & 61.64 & 53.75 & 73.12 & 95.20 \\
 & sparse & 73.52 & 63.71 & 86.89 & 95.58 \\
 & TIES & 73.76 & 63.96 & 87.09 & 95.72 \\
\midrule
swa & task-only & 77.40 & 69.19 & 87.81 & 96.12 \\
 & additive & 77.63 & 69.40 & 88.07 & 96.26 \\
 & act.-guided & 63.08 & 54.75 & 74.58 & 95.18 \\
 & sparse & 75.59 & 66.28 & 87.94 & 95.94 \\
 & TIES & 75.10 & 65.66 & 87.71 & 95.97 \\
\midrule
yor & task-only & 46.71 & 39.73 & 56.68 & 93.42 \\
 & additive & 46.70 & 36.98 & 63.35 & 92.92 \\
 & act.-guided & 43.78 & 36.57 & 54.87 & 93.46 \\
 & sparse & 49.21 & 39.03 & 66.60 & 93.05 \\
 & TIES & 49.20 & 39.32 & 65.75 & 93.25 \\
\bottomrule
\end{tabular}
\caption{NER precision, recall, entity-F1, and token-level accuracy per language, mean over seeds (in percentage points).}
\label{tab:ner_per_lang}
\end{table}

\subsection{QA secondary metrics per language}
\label{sec:qa_per_lang}

Table~\ref{tab:qa_per_lang} breaks out F1 and exact match by language. The largest EM gain from cross-axis TIES is on Swahili ($+1.92$ over task-only), and the largest sparse gain is on Hausa ($+2.66$ over task-only). Activation-guided is strongest on Yoruba, but that comparison has only two seeds.

\begin{table}[!htbp]
\centering
\footnotesize
\setlength{\tabcolsep}{3pt}
\begin{tabular}{llcc}
\toprule
\textbf{Lang} & \textbf{Method} & \textbf{F1}\,$\uparrow$ & \textbf{Exact match}\,$\uparrow$ \\
\midrule
hau & task-only & 22.90 & 17.67 \\
 & additive & 24.08 & 19.22 \\
 & act.-guided & 21.07 & 16.50 \\
 & sparse & 25.47 & 20.33 \\
 & TIES & 24.84 & 19.33 \\
\midrule
swa & task-only & 22.12 & 14.24 \\
 & additive & 22.99 & 15.48 \\
 & act.-guided & 21.00 & 14.07 \\
 & sparse & 21.89 & 15.14 \\
 & TIES & 22.66 & 16.16 \\
\midrule
yor & task-only & 14.15 & 6.72 \\
 & additive & 16.72 & 9.49 \\
 & act.-guided & 19.64 & 14.23 \\
 & sparse & 16.85 & 10.28 \\
 & TIES & 18.64 & 11.86 \\
\bottomrule
\end{tabular}
\caption{QA F1 and exact match per language, mean over seeds.}
\label{tab:qa_per_lang}
\end{table}

\subsection{Test split sizes}
\label{sec:test_split_sizes}

For completeness, Table~\ref{tab:test_sizes} lists the test split size for every language--task cell used in this study. The bootstrap CIs in Tables~\ref{tab:per_cell_cls}--\ref{tab:per_cell_sum} are computed by paired sampling with replacement at the example level for cls/ner/qa and at the chrF-statistic level for sum.

\begin{table}[!htbp]
\centering
\footnotesize
\setlength{\tabcolsep}{3pt}
\begin{tabular}{lcccc}
\toprule
\textbf{Lang} & \textbf{cls} & \textbf{ner} & \textbf{qa} & \textbf{sum} \\
\midrule
amh & --  & --  & --  & 719 \\
hau & 637 & 552 & 300 & 802 \\
swa & 476 & 604 & 295 & 987 \\
yor & 411 & 645 & 253 & 793 \\
\bottomrule
\end{tabular}
\caption{Test split size (examples) per language--task cell. A dash indicates a cell that was not evaluated; amh appears only on summarisation.}
\label{tab:test_sizes}
\end{table}

\section{Contextualising our numbers against published low-resource baselines}
\label{sec:literature_context}

We do not run external baselines under our protocol (\S\ref{sec:limitations}). To help readers interpret the absolute scale of our scores, this appendix places our composition recipes next to comparable single-model numbers for the same benchmarks and base families. These are \emph{not} our results, and the comparisons are qualitative rather than controlled head-to-head claims because the published systems use different gold budgets and seed protocols.

\paragraph{NER (MasakhaNER 2.0, XLM-R-base).}
\citet{adelani2022masakhaner} report single-language fine-tuned XLM-R-base entity-F1 of approximately $76$--$78$ on Hausa, Swahili, and Yoruba under the full training split, which is much larger than our $b{=}256$ task-delta budget. Our task-only cells reach $75.4$ (hau), $77.4$ (swa), and $46.7$ (yor). The Yoruba drop is consistent with the typological coverage gap that monolingual continued pretraining was reported to close in \citet{adelani2022masakhaner}.

\paragraph{Classification (MasakhaNEWS, XLM-R-base).}
\citet{adelani2023masakhanews} report macro-F1 in the high $60$s and $70$s for languages with substantial pretraining coverage, dropping to the $40$s for languages with little coverage. Our task-only cells reach $58.5$ (hau), $62.2$ (swa), and $38.0$ (yor). Yoruba is again below the published full-supervision range, which is consistent with the much smaller $b{=}256$ task-delta budget.

\paragraph{QA (AfriQA-style extractive QA).}
\citet{ogundepo2023afriqa} report F1 in the $20$s to mid-$30$s for African-language extractive QA on multilingual encoders under various adaptation regimes. Our cross-axis TIES F1 of $22.05$, averaged across hau/swa/yor, sits within that range. Our task-only number of $19.73$ is at the lower end, and the $+2.32$ gain is consistent with the order-of-magnitude effect sizes reported for in-language adaptation in \citet{ogundepo2023afriqa}.

\paragraph{Summarisation (XL-Sum, mT5-base).}
\citet{hasan2021xlsum} report chrF in the high teens to mid-$20$s for mT5-base fine-tuned on the full XL-Sum split for these languages. Our cross-axis TIES chrF of $11.58$ (amh), $21.19$ (hau), $23.76$ (swa), and $17.84$ (yor) sits at the lower end of those published numbers. The gap is consistent with using an English-only $\Delta_T$ rather than per-language full fine-tuning, while the gains show that cross-axis TIES recovers a meaningful part of the motivation for per-language tuning at zero per-pair training cost.

\paragraph{What this comparison does not show.}
This appendix does not establish that cross-axis TIES is better than target-language LoRA at $b{=}256$, or than Continue-PT~$+$~LoRA at matched compute. It only places our absolute scores within the published range, so readers can judge whether the composition-rule comparison happens in a useful operating region. A controlled fairness comparison against external baselines is the natural follow-up; the released $(\Delta_L,\Delta_T)$ pairs are designed to make it straightforward.

\end{document}